%% file: main.tex
\documentclass[12pt]{article}
\usepackage[margin=1in]{geometry}
\usepackage{amsmath, amssymb, amsthm}
\usepackage{graphicx}
\usepackage{hyperref}
\usepackage{authblk}
\usepackage[dvipsnames]{xcolor}
\usepackage{float}
\usepackage{tabularx} 
\usepackage{booktabs}

\usepackage[style=numeric, sorting=none]{biblatex}
\begin{document}

\title{Neural Ordinary Differential Equations for Learning and Extrapolating System Dynamics Across Bifurcations} 

\author[1,2]{Eva van Tegelen}
\author[1]{George van Voorn}
\author[2]{Ioannis Athanasiadis}
\author[1]{Peter van Heijster}

\affil[1]{Biometris, Wageningen University and Research}
\affil[2]{Artificial Intelligence Group, Wageningen University and Research}

\date{June 30, 2025}

\maketitle 
\begin{abstract}
\begingroup
\renewcommand{\thefootnote}{}
\footnotetext{Preprint}
\addtocounter{footnote}{-1}
\endgroup
\input{1_abstract}
\end{abstract}

\input{2_introduction}

\input{3_methods}
\input{4_results}
\input{5_discussion}

\printbibliography

\newpage
\appendix
\input{A_tuning}

\end{document}

%% file: 1_abstract.tex
Forecasting system behaviour near and across bifurcations is crucial for identifying potential shifts in dynamical systems. While machine learning has recently been used to learn critical transitions and bifurcation structures from data, most studies remain limited as they exclusively focus on discrete-time methods and local bifurcations. To address these limitations, we use Neural Ordinary Differential Equations which provide a data-driven framework for learning system dynamics. Our results show that Neural Ordinary Differential Equations can recover underlying bifurcation structures directly from time-series data by learning parameter-dependent vector fields. Notably, we demonstrate that Neural Ordinary Differential Equations can forecast bifurcations even beyond the parameter regions represented in the training data. We demonstrate our approach on three test cases: the Lorenz system transitioning from non-chaotic to chaotic behaviour, the Rössler system moving from chaos to period doubling, and a predator-prey model exhibiting collapse via a global bifurcation.

%% file: 2_introduction.tex
\section{Introduction}\label{sec:intro}
Bifurcation theory is a powerful tool to study how and when a system undergoes sudden shifts in its dynamical behaviour. A bifurcation is a point where a small change in the system's parameters or external drivers cause a qualitative shift in its dynamical behaviour. Because such shifts can have significant consequences, ranging from smaller-scale ecosystem collapses \cite{Drake2010-cj,Wang2012-ro,Dakos2019-ef,Willcock2023-dl} to large-scale climate tipping points \cite{Lenton2011-zi,Ashwin2012-ji,Armstrong_McKay2022-dh}, understanding and predicting system behaviour near bifurcation points is crucial \cite{Van_Nes2016-at}. Identifying the types of bifurcations that a real-world system can exhibit provides insights into the system's long-term dynamics and can help guide interventions to prevent undesired outcomes.

A major challenge lies in detecting bifurcations in systems where the underlying processes and governing equations are poorly understood or unknown. In response, data-driven approaches have gained increasing interest. In particular, machine learning is being actively explored as a tool for capturing system dynamics and bifurcations directly from data \cite{Fabiani2021-lz,Rajendra2020-dy}. Deep learning methods, for instance, outperform classical early warning indicators in both sensitivity and specificity when detecting pre-bifurcation signals in time series data\cite{Bury2021-bw,Deb2022-rk,Bury2023-ap}. Parameter-driven reservoir computing has successfully extrapolated the dynamics of bifurcating systems and, in some cases, even predicted behaviour beyond the training data \cite{Lim2020-by,Kong2021-df,Patel2021-ig,Patel2022-ec,Kong2023-lg}. More recently, unsupervised learning techniques, which do not require prior knowledge of the bifurcation parameters, have been shown to infer the parameter values directly from data \cite{panahi2025unsupervised}. Additionally, hybrid modelling, which combines mechanistic modelling with data-driven approximators, such as neural networks, has proven to effectively capture the bifurcation behaviour of diverse systems\cite{Beregi2023-yx,Lee2023-el}. 

Despite these advances, most existing machine learning applications to bifurcations remain limited. Many focus exclusively on discrete machine learning methods \cite{Bury2021-bw,Deb2022-rk,Bury2023-ap,panahi2025unsupervised,Lim2020-by,Kong2021-df,Patel2021-ig,Kong2023-lg}, which rely on discrete-time jumps to capture dynamics. In contrast, many real-world systems evolve smoothly with gradual shifts rather than discrete steps. This mismatch between the discrete assumptions of most machine learning methods and the continuous nature of many systems limits their effectiveness in detecting bifurcations in continuous-time settings. Moreover, much of the focus has been on local bifurcations \cite{Bury2021-bw,Deb2022-rk,Beregi2023-yx,Lee2023-el}, which can be identified from local properties of the system. In contrast, global bifurcations depend on the structure of the large-scale state-space. Because global bifurcations can have significant consequences for system dynamics\cite{van-Voorn2010-sp}, it is important to investigate methods that can infer global system structure from limited data and thereby detect this class of bifurcations.

To address these limitations, we use Neural Ordinary Differential Equations (ODEs); a class of models introduced by Chen et al.\cite{Chen2018-px} that use neural networks to approximate continuous-time dynamics, serving as a data-driven alternative to traditional ODEs. Instead of relying on explicitly defined equations, Neural ODEs can be used to learn a vector field directly from data. This makes them well-suited for modelling complex, nonlinear, or unknown systems\cite{Chen2018-px,Rubanova2019-ev,Wang2022-pk,Coelho2025-aa}. The vector fields learned by Neural ODEs offer a promising approach for uncovering bifurcation structures from data, as they can capture smooth changes in the state-space that may lead to qualitative shifts in system behaviour \cite{Rubanova2019-ev,Kidger2020-es}.

In this work, we demonstrate the potential of Neural ODEs as a tool for extrapolating bifurcation dynamics from data. Although Neural ODEs, and even parameterized modifications\cite{Lee2021-qd}, are established techniques, our approach introduces a novel application: explicitly incorporating a bifurcation parameter as an input to learn bifurcation structures. This specific adaptation enables the model to capture and analyse regime shifts, offering valuable insights into system behaviour. We examine three test cases: the Lorenz system transitioning to chaotic behaviour, the Rössler system moving from chaos to period-doubling and a predator–prey system exhibiting both local and global bifurcations. For each of these systems we investigate the ability of Neural ODEs to extrapolate dynamics across bifurcations. To examine their robustness, we conduct additional experiments on the predator-prey model with limited trajectory data and added measurement noise.

The rest of the paper is organized as follows. In Sec.~ \ref{sec:neuralode}, we give a brief overview on Neural ODEs and how they can be adapted to include bifurcation parameters. In Sec.~\ref{sec:experimentsetup}, we introduce our different test cases and discuss the training procedure. In Sec.~ \ref{sec:results}, we showcase our findings, which will be further discussed in Sec.~\ref{sec:discussion}.

%% file: 3_methods.tex
\section{Neural ODEs with bifurcation parameter} \label{sec:neuralode}
Neural ODEs provide a framework for learning and describing continuous-time dynamics using neural networks\cite{Chen2018-px}. Like classical ODEs they model system evolution through a differential equation:
\begin{equation}
    \frac{dz}{dt}=f_\theta(z,t),
\end{equation}
where $z\in \mathbb{R} ^n$ is a vector of state variables, $f_\theta$ is a function represented by a neural network with $\theta$ denoting its trainable parameters (i.e., weights and biases).

Throughout this paper, we use the term model specifically to refer to the Neural ODE architecture, and model parameters to refer to its internal weights and biases. We will use the terms system and system parameters when referring to a dynamical system and its corresponding equations.

We use an adapted parameterized framework\cite{Lee2021-qd} that allows us to include bifurcation parameters. This way, we can model systems influenced by both internal state variables and external drivers. More specifically, we assume that the dynamics of the system can be described by differential equations of the form:
\begin{equation}
    \frac{dz}{dt}=f_\theta(z,\alpha), \label{eq:neuralode_bif}
\end{equation}
where $\alpha \in \mathbb{R}^m$ are the bifurcation parameters. Note that in our formulation we omit $t$ as input, since we assume for simplicity that our system is autonomous. 

The network takes the current state $z$ and the bifurcation parameter $\alpha$ as inputs and outputs the state’s time derivative. By integrating these derivatives using an ODE solver, Neural ODEs can generate system trajectories from initial conditions, combining the interpretability of traditional ODE modelling with a data-driven approach. 

\begin{figure}[t]
\centering
\includegraphics[width=0.7\linewidth]{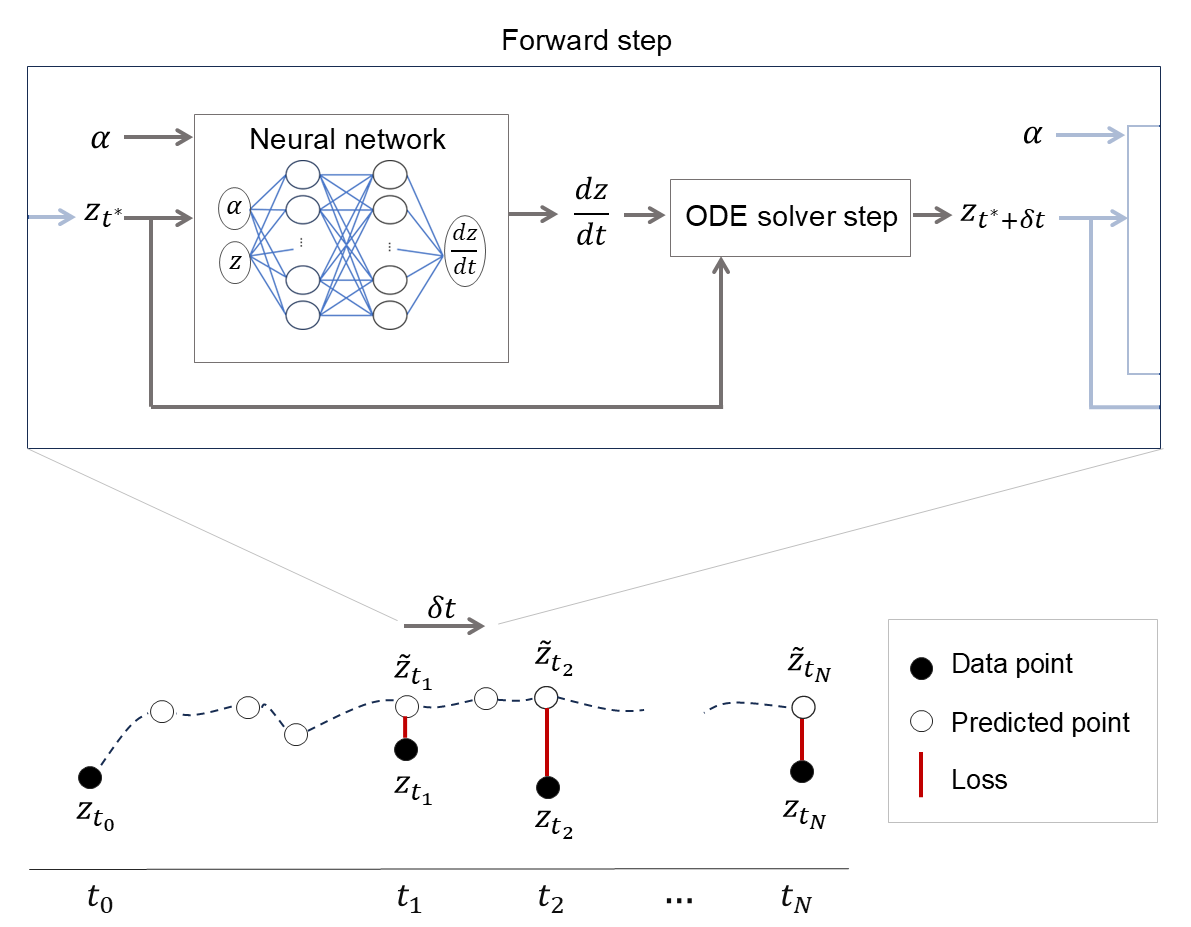}
\caption{\label{fig:NODE} Computing trajectories using a bifurcation parameter-dependent Neural ODE. Top: Zoomed in forward step. The value of the bifurcation parameter and the current state variable are used as input for the neural network. The derivative that is outputted by the neural network together with the previous state is used as input for the ODE solver, which calculates the next state. Bottom: Starting from an initial point ($z_{t_0}$), the Neural ODE iteratively computes the state at next time points using forward steps. The solver can take intermediate time steps between data points. The loss is only calculated at the data points.}
\end{figure}

\begin{figure*}[t]
    \includegraphics[width=\linewidth]{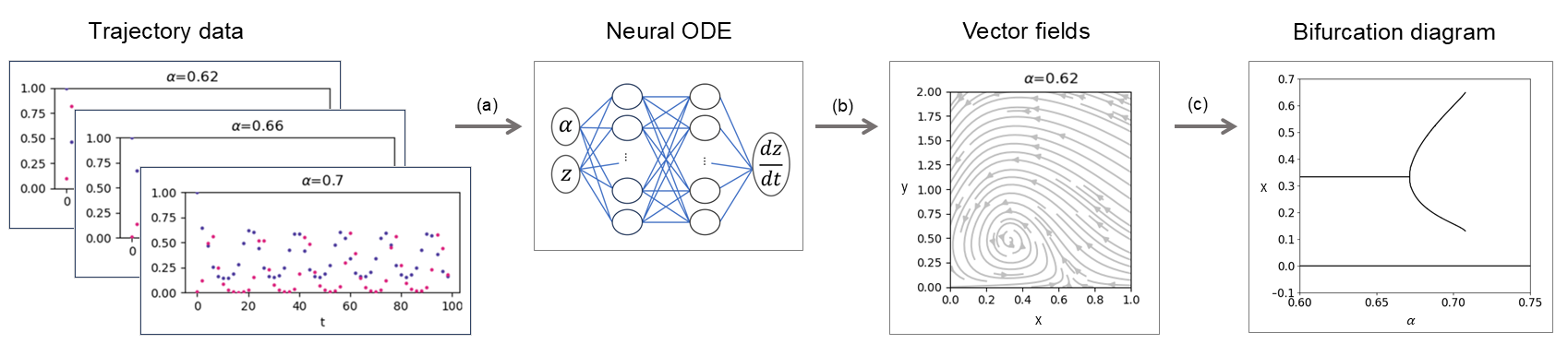}
    \caption{\label{fig:flow} From trajectory data to bifurcation diagram. (a) Trajectory data is used as input to train the Neural ODE. (b) The Neural ODE defines a learned vector field, which can be directly extracted from the model. (c) We determine the bifurcation structure from the vector field (Appendix \ref{Ap:biffigures}).}
\end{figure*}

Representing system dynamics as a vector field allows us to simulate how the system evolves at any point in time, from any initial condition. While any computer implementation must approximate solutions using floating-point arithmetic and discrete solver steps, a Neural ODE defines a numerical approximation to continuous-time vector fields rather than an inherently discrete-time mapping. In contrast, commonly used deep learning methods for time series forecasting, such as Recurrent Neural Networks, learn discrete mappings from one time point to the next, which limits them to fixed time intervals\cite{Rubanova2019-ev,Li2020-bt,Sherstinsky2020-vw}. A continuous representation of dynamics, as used in Neural ODEs, allows us to analyse qualitative system behaviour, such as bifurcations, using approaches similar to those used in the analysis of classical ODEs. 

Neural ODEs compute trajectories by iteratively integrating the learned dynamics forward in time (Fig.~\ref{fig:NODE}). At each forward step, the network receives the current state $z_{t^*}$ and bifurcation parameter $\alpha$ as inputs and outputs the corresponding time derivative of the state $dz/dt$. The outputted derivative together with the current state is integrated using an ODE solver, which then produces the state variables at the next time point $z_{t^*+\delta t}$. This iterative process allows Neural ODEs to generate trajectories that capture the underlying system dynamics. Note that we are not limited to fixed time steps, since the ODE solver can take arbitrarily small or adaptive steps to closely approximate the continuous dynamics of the system.

During training, we optimize the model parameters $\theta$ based on time series data from our system for different values of the bifurcation parameter $\alpha$. During each step in the optimisation process, trajectories are generated from initial conditions, as described above. The loss is computed by comparing the predicted and true states at the observed data points. The model parameters are then updated using backpropagation, which requires computing the gradient of the loss with respect to each of the model parameters. 

In this study, we examine multiple systems to determine whether a Neural ODE can learn bifurcation structures directly from data. We follow the same approach in each case: train a Neural ODE model on time-series data and reconstruct the bifurcation structure from the learned dynamics (Fig.~\ref{fig:flow}). More specifically, we train the model on a particular dynamical regime and investigate whether it can extrapolate to different dynamical behaviours across bifurcations.

\section{Experimental setup}\label{sec:experimentsetup}

\begin{figure}
\centering
\includegraphics[width=0.55\linewidth]{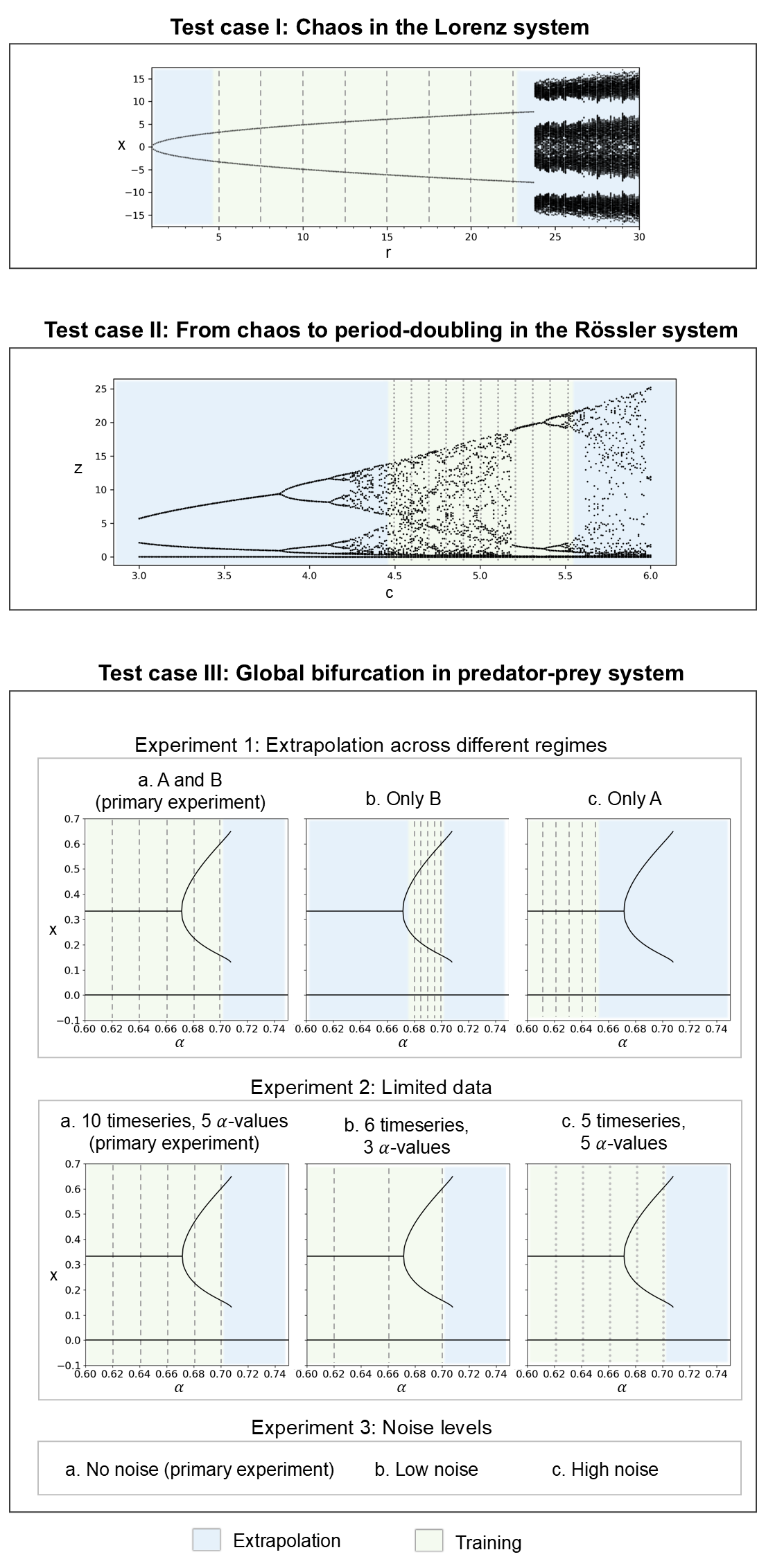}
\caption{\label{fig:allsystems} Bifurcation diagrams and sampling locations of training data for different systems and experiments. Top: Lorenz system. Middle: Rössler system. Bottom: Predator-prey system; Experiment 1: Data is sampled from different regimes; Experiment 2: Amount of data is limited, either by decreasing the number of sampled $\alpha$-values or by reducing the number of initial conditions from two to one; Experiment 3: Different levels of measurement noise are added to the data. Line style indicates amount of initial conditions (striped = two initial conditions, dotted = one initial condition).}
\end{figure}

To test the bifurcation parameter–dependent Neural ODE, we applied it to three test cases. This section describes the different systems and explains the experimental setup that was used. For the first two test cases, the Lorenz and Rössler systems, we conducted a single experiment to showcase the ability of the Neural ODE to infer the bifurcation structure from limited trajectory data. For the third test case, a predator–prey model, we perform additional experiments to investigate how factors such as data size and measurement noise influence the model’s ability to capture the underlying dynamics. Fig.~\ref{fig:allsystems} shows the bifurcation structures of the different systems and highlights the parameter ranges used for training and extrapolation. Additional details on the generation of the synthetic data can be found in \mbox{Appendix~\ref{Ap:datageneration}}.

\subsection*{Test case I: Chaos in the Lorenz system}

For our first test case, we considered the classic Lorenz system:
\begin{equation}
\begin{split}
\dot{x} &= \sigma (y-x), \\
\dot{y} &= r x - y - x z, \\
\dot{z} &= x y - b z,
\end{split}
\end{equation}
with fixed parameters $\sigma = 10$ and $b = 8/3$, and $r$ serving as the bifurcation parameter. Bifurcations and chaos in the Lorenz system have extensively been studied\cite{Sparrow2012-ub}. We focused on the transition from non-chaotic to chaotic behaviour as $r$ is varied. More specifically, we investigated the parameter range ${r\in(1,30)}$, where the Lorenz system becomes chaotic near $r \approx 24.74$. The data consisted of trajectories sampled from the non-chaotic regime at eight different values of the bifurcation parameter $r$ with two distinct initial conditions for each value, yielding a total of sixteen time series.

\subsection*{Test case II: From chaos to period-doubling in the Rössler system}

For our second test case, we considered the Rössler system:
\begin{eqnarray}
\begin{aligned}
\dot{x} &= -y - z, \\
\dot{y} &= x + a y, \\
\dot{z} &= b + z(x - c),
\end{aligned}
\end{eqnarray}
with parameters $a = b = 1/5$ fixed, while treating $c$ as the varying bifurcation parameter. The Rössler system is another well-studied chaotic system \cite{Gardini1985-pp,Sarmah_2013, Rosalie_2016}. Contrary to the previous test case, we trained our model in the chaotic regime and attempted to extrapolate to lower values of $c$ where the system exhibits limit cycles undergoing period-doubling bifurcations. More specifically, we focused on the parameter range ${c\in(3,6)}$. The training data for this test case consisted of ten trajectories sampled at ten different values of the bifurcation parameter $c \in (4.5,5.5)$ with only one initial condition for each value.

\subsection*{Test case III: Global bifurcation in predator-prey system}

For our third test case, we considered the interaction between prey and predator, modelled by the following:
\begin{equation}\label{eq:predatorequation}
\begin{split}
    \dot{x}&=3x(1-x)-xy-\alpha(1-e^{-5x}),\\
    \dot{y}&= -y+3xy,
\end{split}
\end{equation}
where $x$ and $y$ represent the prey and predator populations, respectively, and $\alpha$ serves as the bifurcation parameter. This system of equations extends the classic Lotka–Volterra equations with logistic growth for the prey and an additional nonlinear harvesting term of Ivlev type. This term makes that higher values of the bifurcation parameter $\alpha$, reduce prey growth at higher densities. While this exact formulation has not been studied before, closely related predator–prey models with Ivlev-type responses have been analyzed in the literature\cite{Farris1962-od,Wang2010-zk,Baek2018-iq,Rana2020-fe}, and we provide a detailed analysis of the present system in Appendix~\ref{Ap:predpreybif}.

We limit ourself to the bifurcation parameter range $\alpha \in (0.6, 0.8)$, where the system exhibits two distinct types of bifurcations: a local Hopf bifurcation ($\alpha =0.67$) and a global heteroclinic bifurcation ($\alpha\approx0.71$). We consider three different dynamical regimes. In Regime~A $(0.6,0.67)$ there exists a stable positive steady state, meaning that for a range of initial conditions the predator and prey populations settle into a steady coexistence. Regime B $(0.67,0.71)$ is characterised by limit cycles that persist until Regime C ($\alpha>0.71)$, where the only remaining stable equilibrium is the zero steady state, meaning that both populations eventually go extinct. 

For this test case, we designed a series of simulation experiments to evaluate how well the Neural ODE captures system dynamics and generalises across bifurcations. Similar to the other test cases, a primary dataset was used to showcase that the Neural ODE can infer the full vector field and bifurcation structure from limited trajectory data. In addition, we conducted more targeted experiments to explore how specific factors, such as data size and measurement noise, affect the model's ability to learn the underlying dynamics. Below, we summarize the details of the different experiments.

\begin{itemize}
    \item \textbf{Primary experiment.} The data for the primary experiment consisted of trajectories sampled at five different values of the bifurcation parameter $\alpha$, with two distinct initial conditions for each value, yielding a total of ten time series. Sampling was restricted to regime A and B, enabling us to assess whether the Neural ODE could extrapolate the global bifurcation and collapse in regime C.
    \item \textbf{Experiment 1: Extrapolation across different regimes.} We trained the Neural ODE on trajectories from specific dynamical regimes of the bifurcation diagram to assess its ability to extrapolate to unseen regimes. In one case, the training data was drawn from both regime A and B (the primary dataset). In another, training was restricted to the smaller regime B, which exhibits only limit cycles. Lastly, we trained the model exclusively on data from regime A, where the positive steady state is stable. 
    \item\textbf{Experiment 2: Limited data.}
    We investigated how the amount of training data influences the model performance. To explore this, we created two datasets for which we reduced the available data compared to the primary dataset, either by decreasing the number of sampled $\alpha$-values or by computing trajectories for only a single initial condition.
    \item \textbf{Experiment 3: Noise levels.}
    To study the effect of noisy observations, we introduced measurement noise into the training data and assessed the model performance compared to the primary dataset. The noisy observations were generated by applying the following transformation to our primary dataset:
    \begin{align}
    \begin{split}
    \hat{x}_i&= x_i+\eta_i,\\
    \hat{y}_i&=y_i +\nu_i, \quad\quad {\eta}_i, {\nu}_i \sim \mathcal{N}(0, \sigma^2), 
    \end{split}
    \end{align}
   where $\hat{x}_i$, $\hat{y}_i$ are our noisy data points and $\sigma$ controls the noise level. We considered three cases: no noise $(\sigma=0$; i.e., the primary dataset), low noise $(\sigma=0.05)$, and high noise $(\sigma=0.2)$. Using these values of $\sigma$, the noisy data we generated remained positive for both $x$ and $y$.
\end{itemize}

\subsection*{Neural ODE Architecture and Training Procedure}

For each of the systems we used a similar architecture: a Multilayer Perceptron (MLP) coupled with an ODE solver for time integration. For the Lorenz and Rössler systems we used a fixed-step fourth order Runge-Kutta solver. For the predator–prey system we instead employed the adaptive solver DOPRI5 in order to achieve higher precision. We used the \nolinebreak\textit{torchdiffeq} library \cite{torchdiffeq}, which provides ODE solvers implemented in \textit{PyTorch}. Hyperparameters were selected based on tuning performed on a validation set (See Apppendix~\ref{Ap:tuning} for more details). The number of layers and depth of each layer were varied during the tuning process to explore different model complexities.  Backpropagation can be performed either by differentiating through all intermediate solver steps or by using the adjoint sensitivity method\cite{Chen2018-px}, which reduces memory usage during training. We used standard backpropagation through the internals of the solver, and did not encounter memory or stability issues. 

To guide the training process, we used a mean absolute error (MAE) for the loss function. Compared to the mean squared error (MSE), which disproportionately emphasizes large errors, MAE treats all deviations linearly\cite{Chai2014-ct, Wang2022-pl}. Since MSE would assign greater weight to outliers and might obscure small but meaningful errors during training, MAE is a more suitable objective function for our application. 

It is important to note that for the predator-prey model we additionally added a physics part to the loss function, which enforces that the axes are invariant by penalizing non-zero derivatives of the system along specific axes (See Appendix \ref{Ap:lossfunction}). Since Neural ODEs model derivatives directly, and much of our prior knowledge about physical systems is expressed through differential equations, it is natural to incorporate physics-based information into the loss function. Such physics-informed Neural ODEs have been explored in previous work \cite{O-Leary2022-ey,Sholokhov2023-kz,Li2024-zv}. In our experiments, adding the physics-informed term improved generalisation to unseen parameter ranges by guiding the model training in regions of the vector field not covered by the data. 

%% file: 4_results.tex
\section{Results}\label{sec:results}

\begin{figure}[t]
    \includegraphics[width=\linewidth]{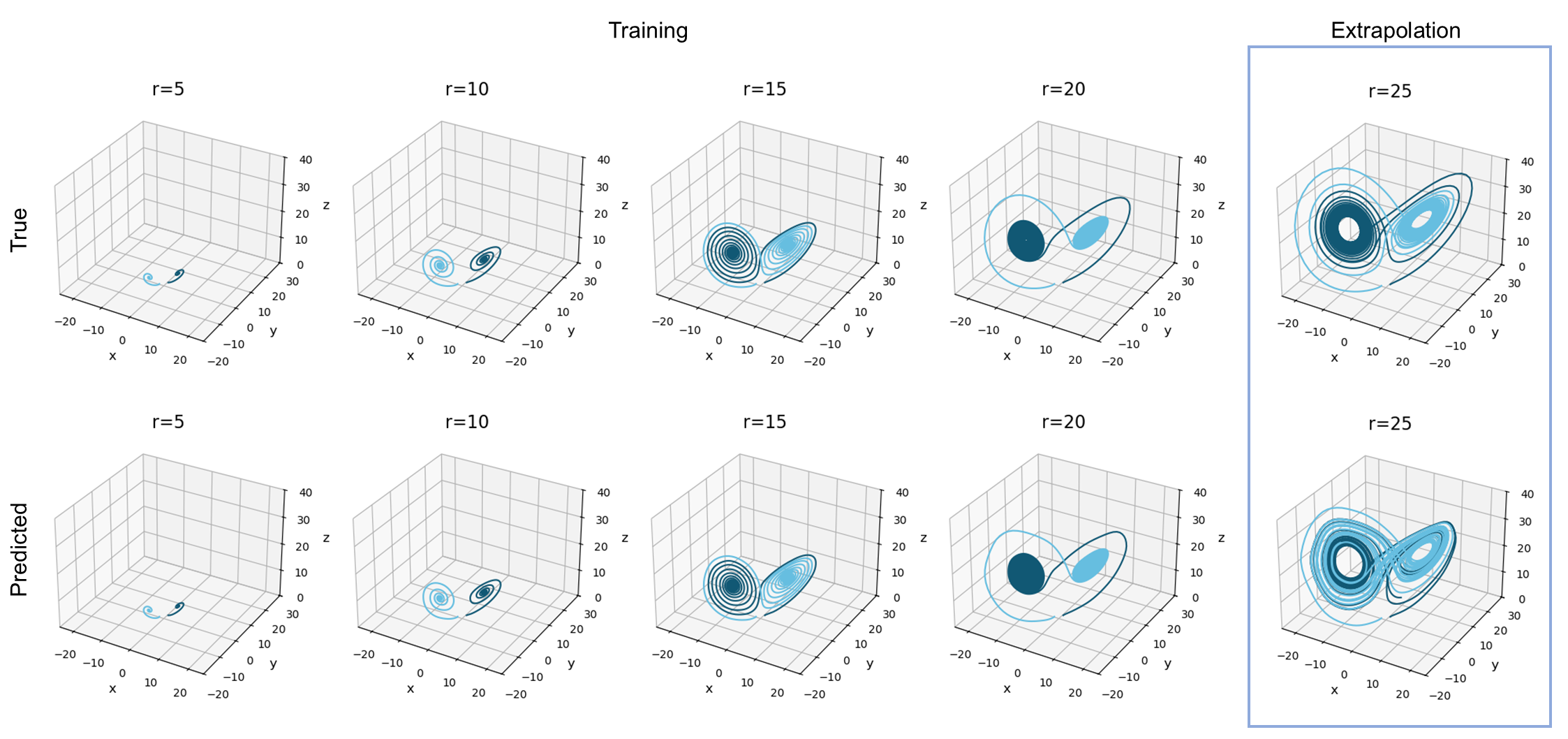}
    \caption{\label{fig:lorenz_traj} Lorenz system trajectories. Top: True system. Bottom: Dynamics learned by the Neural ODE.}
\end{figure}

\begin{figure}[t]
\centering
    \includegraphics[width=0.6\linewidth]{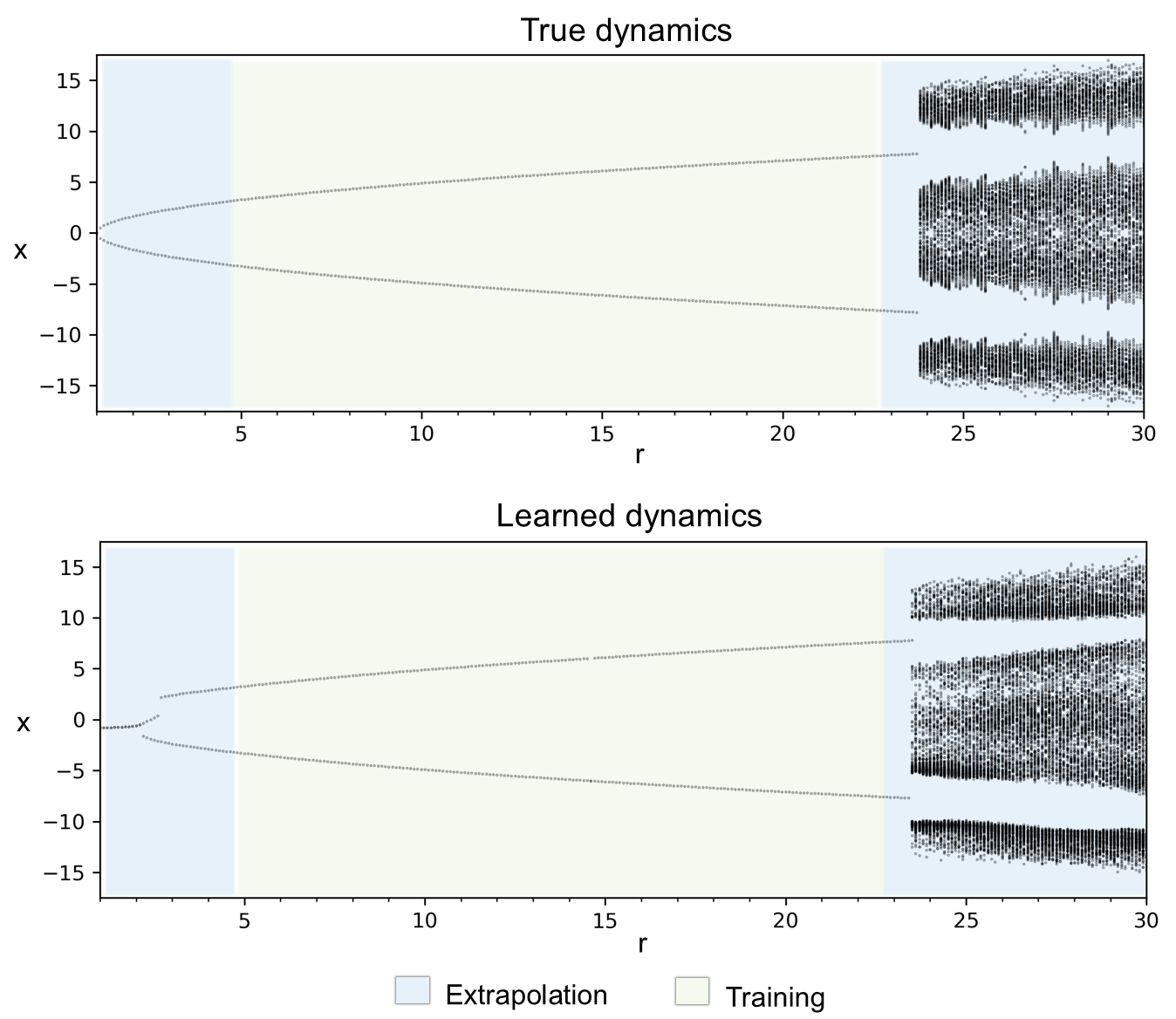}
    \caption{\label{fig:lorenz_bif} Bifurcation diagram for Lorenz system. Top: True system. Bottom: Dynamics learned by the Neural ODE.}
\end{figure}

From the Neural ODE, we are
able to obtain bifurcation structures by simulating the dynamics of the systems. The results are presented in this section. Details on the simulation and visualization process can be found in Appendix~\ref{Ap:biffigures}.

\subsection*{Test case I: Lorenz system}

For the Lorenz system, we trained the Neural ODE on non-chaotic behaviour and evaluated its ability to extrapolate into the chaotic regime. Fig.~\ref{fig:lorenz_traj} shows three-dimensional solution trajectories for multiple values of $r$. The learned trajectories closely match the true trajectories, particularly within the training range. More interestingly, the Neural ODE is able to extrapolate chaotic behaviour beyond the training data. While the extrapolated chaotic trajectories do not exactly replicate the true dynamics, they accurately capture the onset and qualitative features of chaos.

A similar pattern is observed in the bifurcation diagram shown in Fig.~\ref{fig:lorenz_bif}. The Neural ODE successfully learns the two stable fixed-point branches present in the training regime and, more importantly, extrapolates the emergence of chaotic behaviour beyond the training range. The value of $r$ at which chaos first seems to appears in the learned system closely matches that of the true Lorenz system, indicating that the model captures not only the qualitative dynamics but also reasonably approximates the bifurcation point.

To the left of the training regime, however, the extrapolation is less accurate: the two stable fixed-point branches are slightly distorted compared to the true dynamics, reflecting minor limitations in the model’s predictive accuracy farther outside the training range.

\begin{figure*}[h]
    \centering
    \hspace*{-1.5cm}
    \includegraphics[width=1.2\linewidth]
    {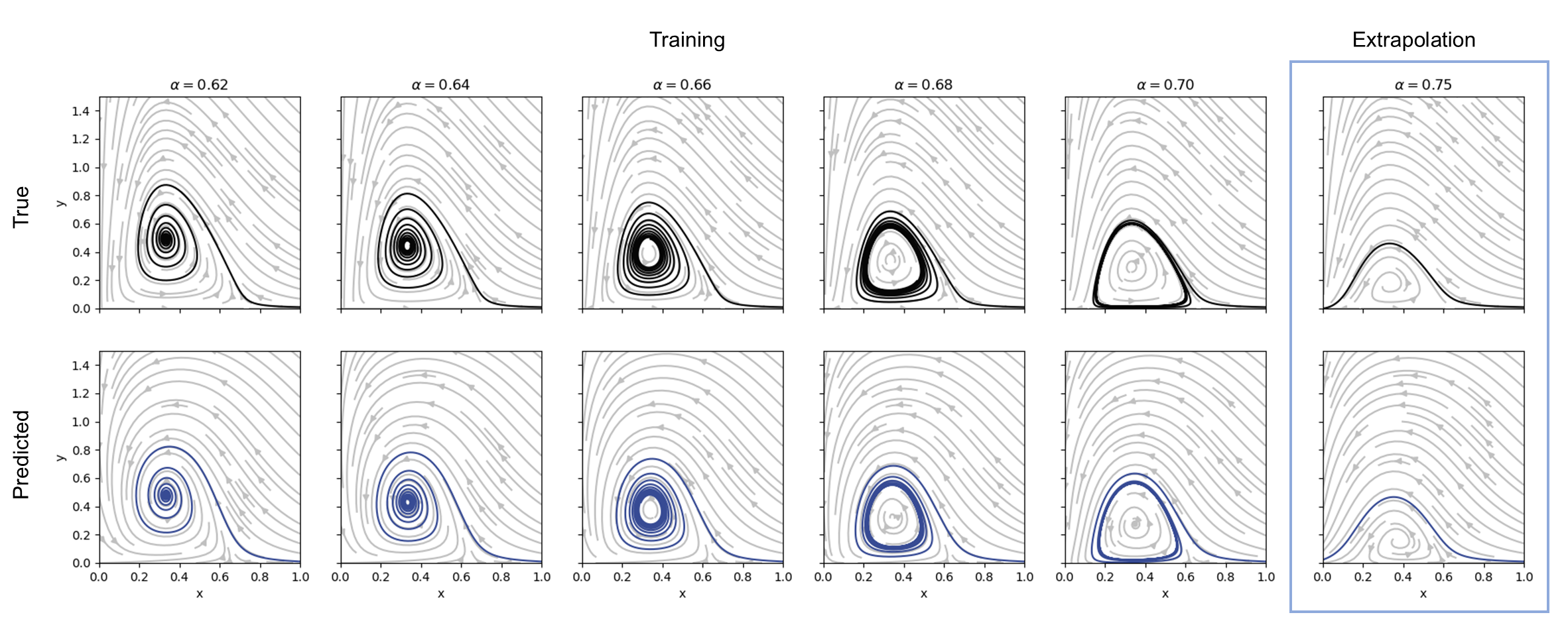}
    \caption{\label{fig:results1} Vector fields describing learned system dynamics. Top: True system. Bottom: Dynamics learned by the Neural ODE. Solid lines are trajectories for the initial condition $(x,y)=(1,0.01)$ to exemplify the system behaviour. The vector field for $\alpha=0.75$ is extrapolated, since this value was not included in the training range.}
\end{figure*}

\subsection*{Test case II: Rössler system}
\begin{figure}[t]
\centering
    \includegraphics[width=0.7\linewidth]{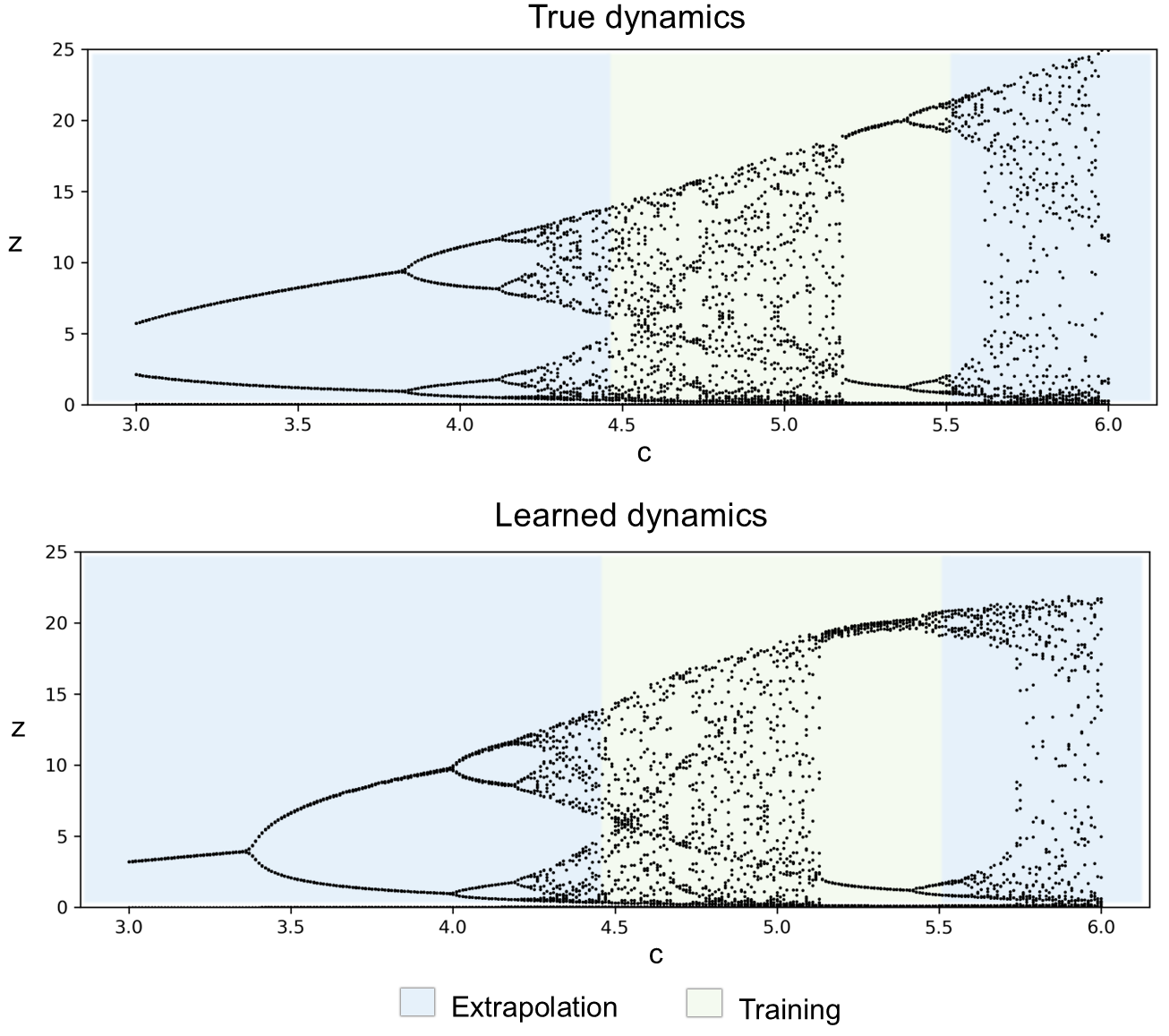}
    \caption{\label{fig:rossler_bif} Bifurcation diagram for Rössler system. Top: True system. Bottom: Dynamics learned by the Neural ODE.}
\end{figure}

For the Rössler system, we trained the Neural ODE on chaotic behaviour and evaluated whether it could learn the underlying dynamics sufficiently to reproduce limit cycles and period-doubling bifurcations occurring at lower values of the bifurcation parameter $c$. Remarkably, despite being trained only on chaotic trajectories, the Neural ODE is able to recover non-chaotic behaviour in the extrapolated regime. The bifurcation diagram in Fig.~\ref{fig:rossler_bif} shows that the model captures not only the chaotic behaviour present in the training data, but also reproduces the limit cycles and period-doubling structure to the left of this regime with reasonable accuracy.

Closer to the training regime, the learned dynamics more closely resemble the true system. However, the exact parameter values at which period-doubling bifurcations occur are slightly shifted compared to the true system. Interestingly, the Neural ODE also predicts an additional period-doubling bifurcation around $c \approx 3.35$, which is not present in the true system. These results highlight both the model’s impressive ability to generalize beyond the training data, but also the subtle limitations of its extrapolation.

\subsection*{Test case III: Predator-prey system}

\begin{figure*}
\centering
\hspace*{-1.7cm}
\includegraphics[width=1.2\linewidth]{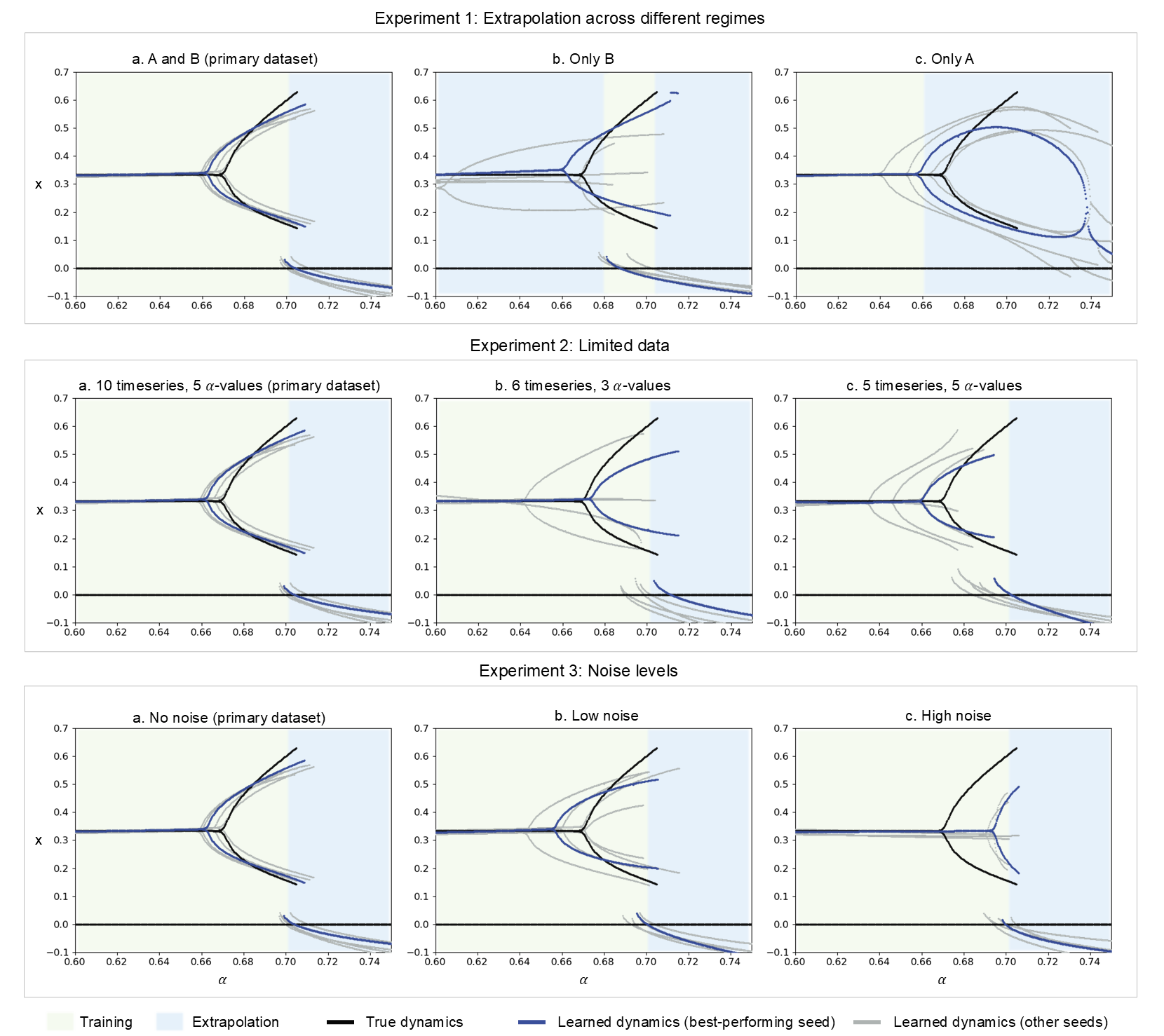}
    \caption{\label{fig:extrapolate} Reconstructed bifurcation diagrams for the predator-prey system for the different experiments.  Each subfigure shows the true dynamics (black), best-performing learned dynamics (blue) and the learned dynamics for other random seeds (grey).}
\end{figure*}

\subsubsection*{Primary experiment}

For the predator-prey system, the model is able to reconstruct vector fields from limited trajectory data, qualitatively capturing the underlying system dynamics, as shown in Fig.~\ref{fig:results1}. Notably, the model not only replicates the behaviour observed in the training trajectories but also extrapolates the onset of a global bifurcation leading to system collapse. Since global bifurcations are characterized by large-scale changes in the state-space that extend beyond local trajectories, this result indicates that the model has captured structural information beyond the training data. A closer examination reveals that the Neural ODE successfully extracts several key dynamic components. In particular, along both the $x$- and $y$-axes, the component of the flow perpendicular to the respective axis remains close to zero. Overall, despite minor deviations, the model accurately reconstructs the qualitative behaviour of the system across different values of $\alpha$.

\subsubsection*{Experiment 1: Extrapolation across different regimes}
We further evaluated the performance of the Neural ODE in different experiments and compared the results to those obtained in the primary experiment. Fig.~\ref{fig:extrapolate} shows the bifurcation diagrams generated by the trained models for each experiment. To account for and showcase the stochastic variability in initialisation and training, the model was trained using five different random seeds. Each of these is displayed in the figure with the best-performing seed, based on MAE relative to the true system, shown in blue. (See Appendix~\ref{Ap:biffigures} for calculation.)

When trained on both the stable coexistence and limit cycle regimes A and B, the model consistently predicts the global bifurcation across all seeds, though the collapse does not reach exactly zero due to the soft nature of the physics-constraint. Training only on the limit cycle regime B yields variable performance: the best seeds capture both the global bifurcation and the transition to a stable focus, while others fail, highlighting the limited coverage of the training data and the model's resulting uncertainty when extrapolating beyond the observed regime. When trained solely on the stable focus regime~A, the model consistently predicts a transition to limit cycles but fails entirely to capture the global bifurcation. This suggests that the training data lacks sufficient information to fully reconstruct key structures of the state-space, such as saddle nodes along the $x$-axis, which play a crucial role in shaping the bifurcation dynamics.

Notably, the zero steady state is absent from the learned dynamics, even in the parameter range it was trained on. This can be explained by the lack of time series in the dataset that converge to zero, making it inherently difficult for the model to learn and reproduce a state that is not represented in the training data.

\subsubsection*{Experiment 2: Limited data}

To assess how the amount of training data influences the model performance,  we trained on datasets with reduced coverage compared to the primary experiment, either by decreasing the number of sampled $\alpha$-values or by limiting the number of initial conditions per $\alpha$-value. Fewer $\alpha$-values led to varied performance across random seeds, even within the training regime; while the best seeds qualitatively reconstructed the bifurcation structure, others were less consistent. Nevertheless, all seeds correctly predicted a global collapse, indicating robustness in identifying critical transitions despite reduced parameter diversity. Limiting initial conditions generally preserved the model’s ability to predict limit cycle dynamics and collapse, though the global bifurcation appeared at a lower $\alpha$-value than in the primary experiment. 

For both types of data reduction, the best-performing random seeds were able to reproduce the qualitative behaviour of the true system. Nonetheless, a visual comparison with results from the primary experiment reveals a noticeable decline in quantitative accuracy (Appendix~\ref{Ap:extrathree}), suggesting that while the overall dynamics remain consistent, the precise locations and occurrences of bifurcations become less reliable with less training data.

\subsubsection*{Experiment 3: Noise levels}

Finally,  we introduced different levels of measurement noise into the training data. At low noise levels, the predicted bifurcation locations and amplitude of limit cycles showed considerable variation across different random seeds. Nevertheless, the best-performing seed still produced a reasonable qualitative approximation of the system dynamics. With higher noise levels, the model’s predictions for the onset of limit cycles became less precise, exhibiting larger deviations compared to the primary, noise-free experiment. Despite this increased uncertainty, the model remained capable of distinguishing between distinct dynamical regimes, successfully identifying both the emergence of limit cycles and the eventual occurrence of the global bifurcation. Further discussion of the quantitative accuracy can be found in Appendix~\ref{Ap:extrathree}.

%% file: 5_discussion.tex
\section{Discussion}\label{sec:discussion}

In this study, we demonstrated that a Neural ODE model can successfully uncover underlying bifurcation structures, training only on limited trajectory data. For the Lorenz system, the model extrapolated chaotic dynamics from non-chaotic training data. For the Rössler system, training on chaotic data enabled qualitatively accurate prediction of non-chaotic behaviour, including period-doubling bifurcations. Experiments on the predator-prey system indicated that Neural ODEs are capable of forecasting global bifurcations beyond the parameter regions represented in the training data. Reducing the amount or quality of training data for this test case decreased consistency across random seeds, while added noise introduced variability but did not remove the model’s ability to recover qualitative regime changes.

While our results highlight the potential of Neural ODEs to learn vector fields directly from trajectory data, they also reveal a key challenge: the quality of extrapolation depends on the information that can be inferred from the training data. When the training data spans a limited part of the state-space and essential aspects, such as saddle nodes, can not be inferred, the model lacks the basic components to reconstruct the full bifurcation structure. This indicates that extrapolation improves when the training data offers more extensive coverage of the system's possible dynamic behaviours.

In real-world systems, however, it is often difficult or impossible to observe the full range of system behaviour, since features like unstable fixed points, are often not represented by the data. In such cases, integrating prior knowledge or mechanistic constraints into the learning process, e.g. through physics-informed neural networks or data augmentation using synthetic data, could enhance a model's ability to generalise to unobserved regions\cite{Raissi2019-wl,Wen2020-fg,Karniadakis2021-sr,von-Rueden2021-ze}. We noted a similar effect in our own work: incorporating a physics-based loss for the predator-prey system led to improved model performance, since it gave the model valuable information about the state-space.

In our paper, we kept the bifurcation parameters constant within each trajectory to enable more interpretable analysis of the model’s extrapolation behaviour. Allowing bifurcation parameters to vary dynamically would effectively place trajectories in a higher-dimensional state-space, making the dynamics more complex to visualise and interpret. Starting with constant parameters therefore provided a natural and informative first step before extending the approach to scenarios involving time-dependent bifurcation parameters. While our experimental setup was deliberately simplified for clarity, the method itself is not limited to this scenario. It is broadly applicable to any continuous dynamical system for which trajectory data is available. In particular, the Neural ODE framework can incorporate dynamically-changing parameters with minimal modification, making it well-suited for real-world systems with external drivers that vary over time.

In this study, we used a basic Neural ODE architecture to model the system dynamics. While this basic formulation already demonstrated strong potential, a range of extensions to the Neural ODE framework have been proposed \cite{Poli2019-zg,Zhong2019-sa,Kidger2020-es,Dandekar2020-ri,Yi2023-pm,Fronk2023-cr}, including Augmented Neural ODEs \cite{Dupont2019-lj}, which may offer improved accuracy in capturing complex dynamical behaviours or greater robustness in the presence of noise. Compared to discrete-time approaches such as Recurrent Neural Networks (RNNs) or Long Short-term Memory Networks (LSTMs), Neural Ordinary Differential Equations naturally operate in continuous time and directly approximate the underlying vector field, which can be advantageous when analyzing bifurcations. A promising direction for future work would be to systematically compare the performance of standard Neural ODEs, their extensions, and discrete-time machine learning methods in tasks such as forecasting global bifurcations and reconstructing bifurcation diagrams from data. While such comparisons are not straight-forward, especially since discrete-time models typically lack access to the underlying vector field and operate at fixed time steps, they could provide valuable insight into the trade-offs between the continuous and discrete frameworks in terms of accuracy, interpretability, and computational cost.

While the above comparison focusses on different machine learning models for forecasting and reconstructing bifurcations, another valuable perspective is to examine how the proposed approach compares to, and could potentially be integrated with, alternative frameworks for predicting critical transitions. A common approach to predicting critical transitions, specifically local bifurcations, is through early warning signals based on slowed recovery after small perturbations \cite{Scheffer2009-jb,Scheffer2012-be,Lenton2012-xx}. Unfortunately, these signals are not observed before all tipping points, including global bifurcations \cite{Ditlevsen2010-tb,Hastings2010-xv,Boerlijst2013-ud}. Furthermore, because these signals do not model or capture the system’s dynamics, they cannot be used to extrapolate behaviour beyond the bifurcation point \cite{Kefi2013-kg}. This limitation points to a potential integration with Neural ODEs. Since early warning signals typically rely on process noise to reveal subtle changes in stability, introducing similar noise-driven variability into the training data could help the Neural ODE detect potential stability changes in trajectory data more reliably. In future work, these signals could potentially be treated as an additional input to improve the model’s ability to anticipate and reconstruct complex bifurcations.

In conclusion, this work demonstrates that Neural ODEs may be used not only to capture system dynamics from trajectory data, but also to recover underlying bifurcation structures, such as heteroclinic bifurcations, chaos and period-doubling. By explicitly including the bifurcation parameter and learning the full vector field, our approach offers a flexible and interpretable framework for studying critical transitions in dynamical systems. These results motivate future applications to more realistic systems, including those with multiple or time-varying parameters and unknown environmental interactions.

\section*{Acknowledgements}
The authors acknowledge Wageningen University and Research for their investment program Data Science and Artificial Intelligence.

\section*{Data Availability Statement}
All code and data that support the findings of this study are available on Github: 
\newline
\href{https://github.com/evantegelen/neural-bifurcations}{https://github.com/evantegelen/neural-bifurcations}.

%% file: A_tuning.tex
\appendix 
\section{System analysis predator-prey}\label{Ap:predpreybif}
\begin{figure}[h]
\centering
\includegraphics[width=0.7\linewidth]{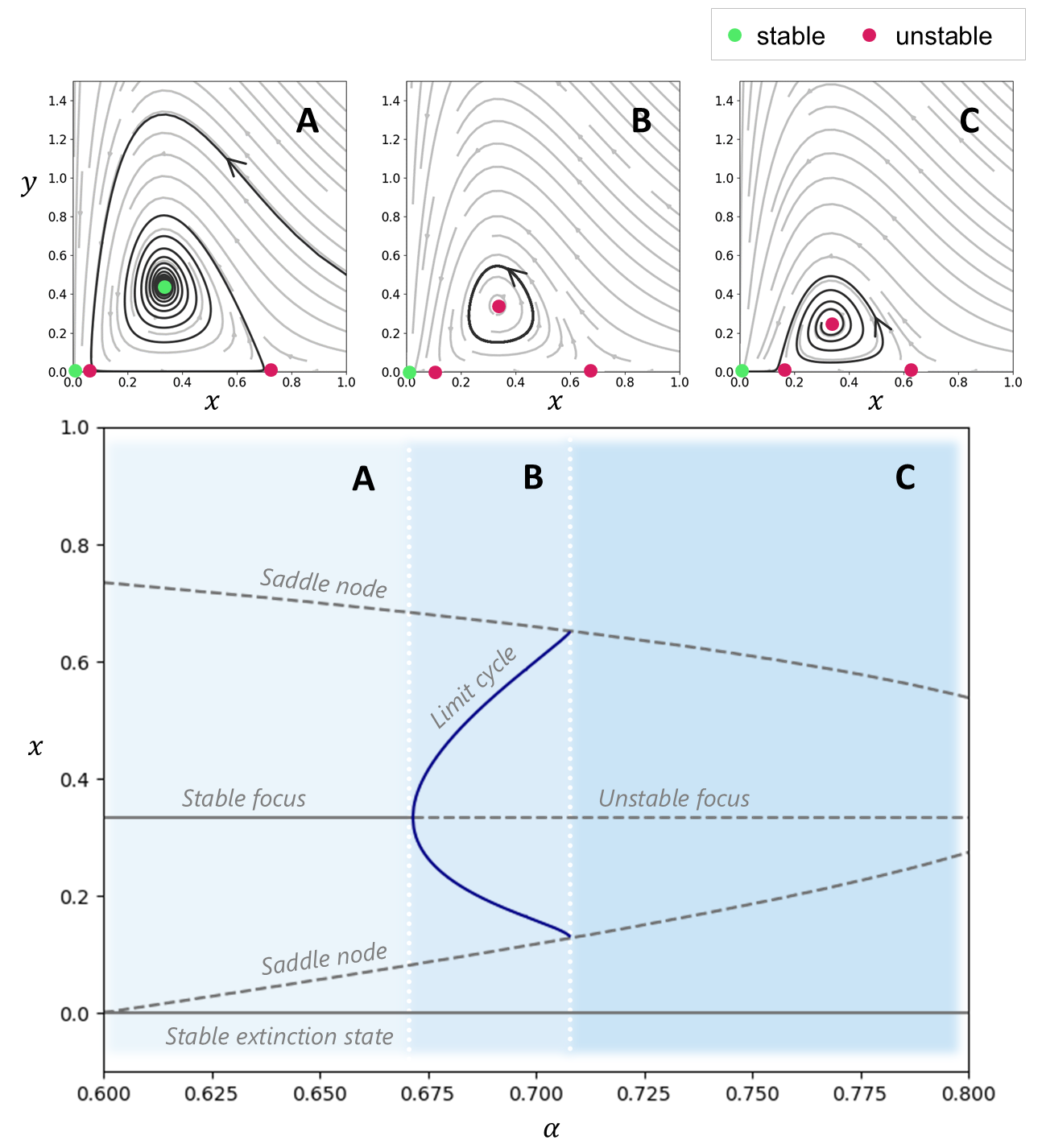}
\caption{\label{fig:predprey} Bifurcation structure of predator-prey system. Top: Vector fields showcasing the distinct behaviours in each of the regimes. A: stable focus, B: stable limit cycle, C: unstable focus. Bottom: Bifurcation diagram demonstrating dynamical behaviour for different values of $\alpha$. A $\rightarrow$ B: Hopf bifurcation. B$\rightarrow$C: heteroclinic bifurcation. Limit cycles are visualised by plotting the minimum and maximum value. } 
\end{figure}

In this appendix we provide more details on the system dynamics of the predator-prey system. Fig.~\ref{fig:predprey} illustrates the bifurcation structure of the system, highlighting the different dynamical regimes. Note that we limit ourself to the bifurcation parameter range $\alpha \in (0.6, 0.8)$, since the zero steady state is only stable for $\alpha > 0.6$. Within this range the system exhibits three distinct dynamical regimes, which allows us to analyse if the model is able to extrapolate transitions between qualitatively different behaviours. The system has four steady states: the zero steady state, two saddle nodes on the $x$-axis and a coexistence steady state. 

In Regime~A the coexistence steady state is a stable focus, meaning that for a range of initial conditions the predator and prey populations settle into a steady coexistence. At $\alpha =0.67$, the system undergoes a Hopf bifurcation, causing the coexistence steady state to lose stability, giving rise to stable limit cycles, which characterise Regime B. These limit cycles persist until $\alpha\approx0.71$, where the system undergoes a global heteroclinic bifurcation; the limit cycle collides with the two saddle nodes, leading to a collapse. In Regime C ($\alpha>0.71)$, the only remaining stable equilibrium is the zero steady state, meaning that both populations eventually go extinct, marking a critical transition point, that has previously been interpreted as an ecological phenomenon referred to as over-exploitation\cite{Van_Voorn2007-lb}.

\section{Data generation}\label{Ap:datageneration}

In this section, we provide additional details about the data used to train the models for the various systems. Table~\ref{tab:datasetup} summarizes the initial conditions and bifurcation parameter values that were used to generate the time series data. For the Lorenz and Rössler systems, we normalised the data using standard normalisation, whereas for the predator–prey model, the data was left unnormalised.

\begin{table*}
\centering
\caption{\label{tab:datasetup} Overview of experimental setup for different systems. Bifurcation parameter values, initial conditions, and noise levels for the different datasets. The values of the bifurcation parameter that were used as the validation set are indicated in grey. Predator-prey system: differences with respect to the primary experiment are indicated in cursive.}
\hspace*{-1.6cm}
\begin{tabular}{llll}
\\[10pt]
\textbf{} & \textbf{Bifurcation parameter values} & \textbf{Initial conditions} & \textbf{Noise levels} \\[8pt]
\hline \hline
\textbf{Test case I: Lorenz} &  $r$ & $(x,y,z)$   \\
\hline
& 5.0, 7.5, 10, 12.5, 15,  & (1,1,1),(-1,-1,1)\\
&17.5, 20, \textcolor{gray}{22.5}\\[10pt] \hline
\hline
\textbf{Test case II: Rössler} &  $c$ & $(x,y,z)$   \\
\hline
& \textcolor{gray}{4.5}, 4.6, 4.7, 4.8, 4.9& (-4, -4, 0)\\
& 5.0, 5.1, 5.2, 5.3, 5.4 \\[10pt] \hline
\hline
\textbf{Test case III: Predator-prey} & $\alpha$ & $(x,y)$ & $\sigma$ \\
\hline
\multicolumn{4}{l}{$\quad$\textbf{Primary experiment}} \\
 & 0.62, 0.64, 0.66, 0.68, \textcolor{gray}{0.7} & (1,0.01),(1,0.1) & - \\
\hline
\multicolumn{4}{l}{$\quad$\textbf{Experiment 1}} \\
$\quad$A and B & 0.62, 0.64, 0.66, 0.68, 0.7 & (1,0.01),(1,0.1) & - \\
$\quad$Only B & \textit{0.68, 0.685, 0.69, 0.695, 0.7} & (1,0.01),(1,0.1) & - \\
$\quad$Only A & \textit{0.61, 0.62, 0.63, 0.64, 0.65} & (1,0.01),(1,0.1) & - \\
\hline
\multicolumn{4}{l}{$\quad$\textbf{Experiment 2}} \\
$\quad$10 time series, 5 $\alpha$-values & 0.62, 0.64, 0.66, 0.68, 0.7 & (1,0.01),(1,0.1) & - \\
$\quad$6 time series, 3 $\alpha$-values & \textit{0.62, 0.66, 0.7} & (1,0.01),(1,0.1) & - \\
$\quad$5 time series, 5 $\alpha$-values & 0.62, 0.64, 0.66, 0.68, 0.7 & \textit{(1,0.1)} & - \\
\hline
\multicolumn{4}{l}{$\quad$\textbf{Experiment 3}} \\
$\quad$No noise & 0.62, 0.64, 0.66, 0.68, 0.7 & (1,0.01),(1,0.1) & - \\
$\quad$Low noise & 0.62, 0.64, 0.66, 0.68, 0.7 & (1,0.01),(1,0.1) &\textit{0.05} \\
$\quad$High noise & 0.62, 0.64, 0.66, 0.68, 0.7 & (1,0.01),(1,0.1) & \textit{0.2} \\
\end{tabular}

\end{table*}

\subsection*{Test case I: Lorenz}
We used numerical integration of the Lorenz system to simulate the trajectories for the given values of the bifurcation parameters and initial conditions. Starting from time $t=0$, we sampled 100 time points with time step $dt=0.05$. 

\subsection*{Test case II: Rössler}
For the Rössler system, we similarly used numerical integration to simulate the trajectories. We wanted the system already to be converged to an attractor, so we used a warm-up time of $t=475$ and them sampled 200 time points with a time step $dt=0.125$.

\subsection*{Test case III: Predator-prey}

For the predator-prey system, for each configuration illustrated in Table~\ref{tab:datasetup}, we used numerical integration to simulate the solution trajectories and obtained time series data by sampling at discrete time steps $t = {0, 1, 2, \dots, 100}$. Note that in each of the experiments one of the datasets overlaps with the primary experiment. This was done for easy comparison.

\begin{figure}
\centering
\includegraphics[width=0.6\linewidth]{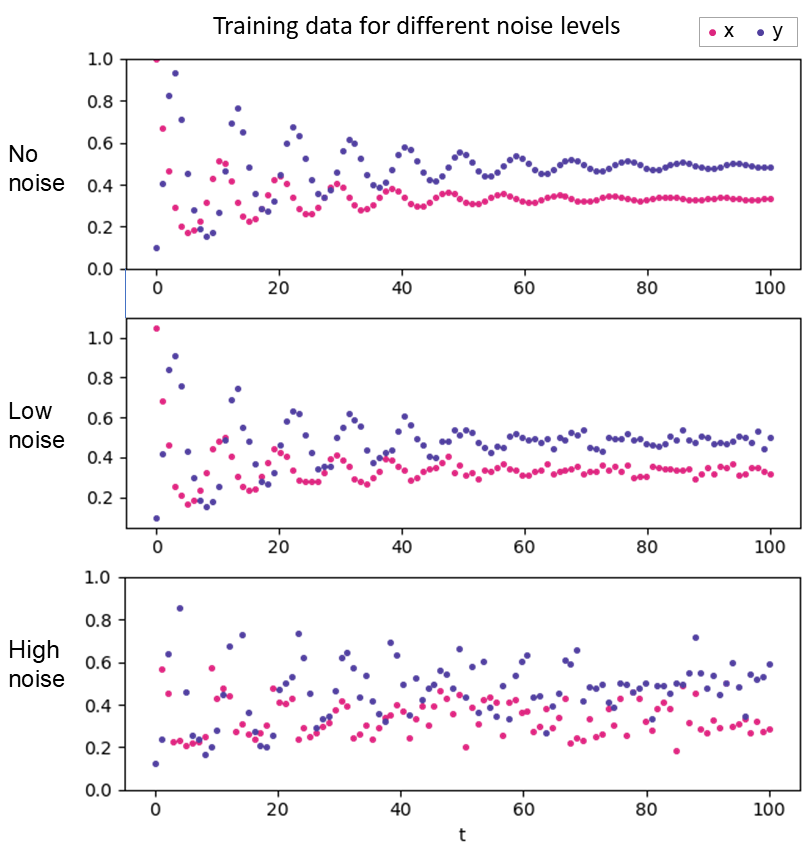}
\caption{\label{fig:noise_data} Time series data for~$\alpha=0.62$ for with different noise levels.}
\end{figure}

Fig.~\ref{fig:noise_data} provides an example of the trajectory data for a specific value of $\alpha$. To highlight the impact of the added measurement noise, the figure also displays the corresponding time series under different noise levels,  illustrating how the original dynamics become increasingly distorted as the noise level increases

.

\section{Training and tuning procedure}

\subsection{Loss function}\label{Ap:lossfunction}
The loss function for the Lorenz and Rössler test cases was defined using the MAE:
\begin{equation}
\mathcal{L} =  \frac{1}{N} \sum_{i=1}^{N} {\| \mathbf{z}_i-\tilde{\mathbf{z}}_i \|}_1,
\end{equation}
where $\mathbf{z}_i = [x_i,y_i]$ are our training data points, $\tilde{\mathbf{z}}_i=[\tilde{x}_i,\tilde{y}_i]$ are our predicted state variables and $N$ is the number of data points.
For the predator-prey system, to guide the training process, we combined a MAE with a physics-informed term to obtain our loss function:
\begin{eqnarray}
\mathcal{L} = && \frac{1}{N} \sum_{i=1}^{N} {\| \mathbf{z}_i-\tilde{\mathbf{z}}_i \|}_1  \label{eq:lossfunction}\\ 
&& +\lambda \left[
    \frac{1}{M} \sum_{j=1}^{M} \left| \frac{d\tilde{x}}{dt}(0,y_j)  \right| +\right. \nonumber \left.\frac{1}{M} \sum_{j=1}^{M} \left|  \frac{d\tilde{y}}{dt} (x_j,0) \right|\right],
\end{eqnarray}
 where $M$ indicates the number of points on which we evaluate the physics-loss. The physics part of the loss function enforces that the axes are invariant by penalizing non-zero derivatives of the system along specific axes. In our experiments, adding the physics-informed term improved generalisation to unseen parameter ranges by guiding the model training in regions of the vector field not covered by the data. The hyperparameter $\lambda$ is the physics-loss weight and determines how much the physics-loss contributes to the total loss. Note that the physics-informed term is not a hard constraint, but a soft constraint driving the derivatives on the axes towards zero.

\subsection{Batching}\label{Ap:batching}
During training, we employed batching to improve convergence and efficiency. All time series were segmented into non-overlapping blocks of a fixed number of time steps, defined by the batch length. A training batch was then formed by randomly selecting a number of these blocks (equal to the batch size). For each batch, the loss was computed and the network parameters were updated accordingly. At every epoch, new batch combinations were generated by randomly re-sampling blocks from the full training set, ensuring diverse training input across epochs.

\subsection{Hyperparameter tuning} \label{Ap:tuning}
To perform hyperparameter tuning, we split our data for each of the systems into a training and validation set, ensuring that the validation set contained a value of the bifurcation parameter outside the range seen during training to specifically test the model’s ability to extrapolate. The trajectories that were used as validation are highlighted in Table~\ref{tab:datasetup} in grey. We conduct a random grid search (Lorenz and Rössler) or full grid search (predator-prey) and for each selected parameter combination we evaluated the model across three different random seeds to account for stochastic variability in model initialization and training. For each trial, we computed the following MAE on the validation set: 
\begin{equation}
  \mathit{MAE}_{val}  = \frac{1}{N_{val}} \sum_{i=1}^{N_{val}} {\| \mathbf{z}_i-\tilde{\mathbf{z}}_i \|}_1,
\end{equation}
where $\mathbf{z}_i = [x_i,y_i]$ are our validation data points,  $\tilde{\mathbf{z}}_i$ our predicted state variables, and $N_{val}$ is the number of validation data points. The investigated and selected hyperparameters for each of the systems are listed in Table~\ref{tab:hyperparams}.

For the predator-prey system multiple experiments were done with different datasets, however we tuned our model parameters only once for the primary experiment. The hyperparameters remained fixed across all other experiments. 
In addition to the tuned hyperparameters, several other parameters were held fixed throughout all experiments. These are listed in Table~\ref{tab:fixed_params}.

\begin{table}[ht]
\centering
\caption{Hyperparameters considered during tuning, along with their tested values. Best performing hyperparameter combinations are indicated in bold (lowest average MAE over three different random seeds).}
\label{tab:hyperparams}
\begin{tabular}{ll}
\\[7pt]
\textbf{Hyperparameter} & \textbf{Values}\\[5pt]
\hline \hline
\multicolumn{2}{l}{\textbf{Test case I: Lorenz}}\\
\hline
Learning rate &0.0005, \textbf{0.001}, 0.005 \\
Number of layers & 1,\textbf{ 2,} 3 \\
Layer depth & 16, 32, \textbf{64}\\
Batch size & 7, \textbf{14}\\
Sequence length & 10, \textbf{20}\\[9pt]
\hline \hline
\multicolumn{2}{l}{\textbf{Test case II: Rössler}}\\
\hline
Number of layers & 1, \textbf{2}, 3 \\
Layer depth & 16, 32, \textbf{64}\\
Batch size & 5, \textbf{10}, 25\\
Sequence length & 10, 20, \textbf{40}\\[9pt]
\hline \hline
\multicolumn{2}{l}{\textbf{Test case III: Predator-prey}}\\
\hline
Learning rate &\textbf{0.0001}, 0.001 \\
$\lambda$ (physics loss weight) & 0.1, \textbf{0.01}, 0.001 \\
Number of layers & 1, 2, \textbf{3} \\
Layer depth & 32, \textbf{64} \\
\end{tabular}
\end{table}

\begin{table}[ht]
\centering
\caption{Fixed hyperparameters. These were not varied during tuning.}
\label{tab:fixed_params}
\begin{tabular}{ll}
\\[7pt]
\textbf{Hyperparameter} & \textbf{Value}\\[5pt]
\hline \hline
\multicolumn{2}{l}{\textbf{Test case I: Lorenz}}\\ \hline
ODE solver step  & $0.05$ \\
Number of epochs & 10000 (during tuning)\\
&20000 (final experiments)
\\[9pt]
\hline \hline
\multicolumn{2}{l}{\textbf{Test case II: Rössler}}\\
\hline
Learning rate & 0.001\\[9pt]
ODE solver step  & $0.1$ \\
Number of epochs & 7500 (during tuning)\\
&15000 (final experiments)
\\[9pt]
\hline\hline\multicolumn{2}{l}{\textbf{Test case III: Predator-prey}}\\
\hline
ODE solver tolerance ($r_{\text{tol}}$) & $1 \times 10^{-5}$ \\
Batch size & 5 \\
Sequence length & 20 \\
Number of epochs & 5000 (during tuning)\\
&10000 (final experiments) \\
\hline
\end{tabular}
\end{table}

\section{Generating bifurcation diagrams from Neural ODEs}\label{Ap:biffigures}
To investigate the dynamical behaviour of our trained models, we used a simulation-based approach to generate the bifurcation diagrams that were shown in Fig.\ref{fig:lorenz_bif}, Fig.\ref{fig:rossler_bif} and Fig.\ref{fig:extrapolate}. These simulations were performed after training the models using the learned vector fields. We will shortly go into how the bifurcation diagrams were constructed for each of the systems.

\subsection*{Test case I: Lorenz}
For 290 different values of the bifurcation parameter $r\in(1,30)$ we ran the model until $t=250$ from two initial conditions: $(x,y,z)=(1,1,1)\nolinebreak$ and $(x,y,z)=(-1,-1,1)\nolinebreak$. After a warm-up time, which we take to be $t=200$, we plot the $x$-value of the local minima and maxima in our trajectories. For the visualisation in Fig. \ref{fig:lorenz_bif} we used the random seed with the lowest training MAE.

\subsection*{Test case II: Rössler}
For 300 different values of the bifurcation parameter $c\in(3,6)$ we ran the model until $t=1000$ from one initial condition: $(x,y,z)=(-4,-4,0)\nolinebreak$. After a warm-up time of $t=900$, we plot the $z$-value of the local minima and maxima in our trajectories. For the visualisation in Fig.~\ref{fig:rossler_bif} we used the random seed with the lowest training MAE.

\subsection*{Test case III: Predator-prey}

For 500 different values of the bifurcation parameter $\alpha$, we ran each model until $t=2000$ from two distinct initial conditions: $(x,y)=(0,0)\nolinebreak$ and $(x,y)=(1,0.01)\nolinebreak$. From the final 500 time points of each simulation, we extracted the minimum and maximum value for the $x$-component, which serve as indicators of the system’s long-term behaviour. If the model converges to a fixed point, these values coincide; in the case of a limit cycle, they reflect the oscillation amplitude in the $x$-direction. To ensure a fair comparison, we applied the same procedure to the true system using identical initial conditions and parameter values. We restrict our bifurcation diagrams for the $x$-component, as it captures the most informative aspects of the system dynamics.

To determine the best performing seed we used MAE as the error metric on the simulated converged behaviour. Specifically, from the simulations described above we calculated the bifurcation error $\text{MAE}_{\mathit{bif}}$ using the minimum and maximum values for the initial condition $(x,y)=(1,0.01)$ as follows: 

\begin{eqnarray}
\begin{aligned}
  &\text{MAE}_{\mathit{max}} = \frac{1}{N_a} \sum_{\alpha=1}^{N_a} \left| x_{\alpha,\mathit{max}} - \tilde{x}_{\alpha,\mathit{max}} \right|,\\
  &\text{MAE}_{\mathit{min}} = \frac{1}{N_a} \sum_{\alpha=1}^{N_a} \left| x_{\alpha,\mathit{min}} - \tilde{x}_{\alpha,\mathit{min}} \right|,\\
  &\text{MAE}_{\mathit{bif}} = \text{MAE}_{\mathit{max}}+\text{MAE}_{\mathit{min}},
\end{aligned}\label{eq:weirdmae}
\end{eqnarray}

where $N_a$ is the number of $\alpha$-values that was used for simulation. The $\text{MAE}_{\mathit{bif}}$ was used to select the best performing seed for each of the experiments (shown in Fig.~\ref{fig:extrapolate}). 

\section{Extra results on test case III}\label{Ap:extrathree}
Table~\ref{tab:errordata} shows the mean and standard deviation of the MAE (calculated as in Eq.~(\ref{eq:weirdmae})) for each of the experiments. Note that this error metric differs from the ones used for training and validation. 
\begin{table}[]
\caption{Mean $\text{MAE}_{\mathit{bif}}$ over different seeds for the different experiments. Right column is the standard deviation on the $\text{MAE}_{\mathit{bif}}$.\label{tab:errordata}}
\centering
\begin{tabular}{llccc}
\\[3pt]
$\text{MAE}_{\mathit{bif}}$& Mean &$\pm$ & Standard deviation\\
\hline 
\textbf{Primary experiment}
& 0.036 &$\pm$ & 0.004 \\
\hline
\textbf{Experiment 1} \\
A and B & 0.036 &$\pm$ & 0.004 \\
Only B & 0.074 &$\pm$& 0.009 \\
Only A & 0.092 &$\pm$& 0.007 \\
\hline
\textbf{Experiment 2}\\
10 time series, 5 $\alpha$-values & 0.036 &$\pm$ & 0.004 \\
6 time series, 3 $\alpha$-values & 0.072 &$\pm$& 0.014 \\
5 time series, 5 $\alpha$-values & 0.087 &$\pm$& 0.025 \\
20 time series, 10 $\alpha$-values & 0.033 &$\pm$& 0.006
 \\
\hline
\textbf{Experiment 3}\\
No noise & 0.036 &$\pm$ & 0.004\\
Low noise & 0.054 & $\pm$&0.011 \\
High noise & 0.059 &$\pm$& 0.005 \\
\end{tabular}

\end{table}

Table ~\ref{tab:errordata} shows quantitatively how the model performed for the different experiments. Similar conclusions can be drawn as from the visual representation in Fig.~\ref{fig:extrapolate}. For Experiment 1, when comparing to the primary experiment, only training on regime B decreased the overall accuracy and increased the variation between the different seeds. The accuracy decreased even further when only training on regime A.

For Experiment 2, it can be seen that decreasing the amount of data does impact the accuracy quite a bit. There is also a clear increase in the standard deviation, indicating that the differences in performance between the seeds increases. This coincides with what we saw in Sec.~\ref{sec:results}; some random seeds were still able to reconstruct the bifurcation structure qualitatively well, but others failed. The table also shows an additional dataset, where the amount of data was increased, of which the results are discussed below. Interestingly, doubling the amount of time series data does not lead to a big decrease in the MAE.

For Experiment 3, the table shows that adding measurement noise to the data indeed impacts the performance of the model. With low noise levels the accuracy goes down compared to the primary experiment, but also the performance varies more between seeds. For high levels of noise there is less varied performance between the seeds, but the overall accuracy is lower. 

\begin{figure}[h]
\centering
\includegraphics[width=0.8\linewidth]{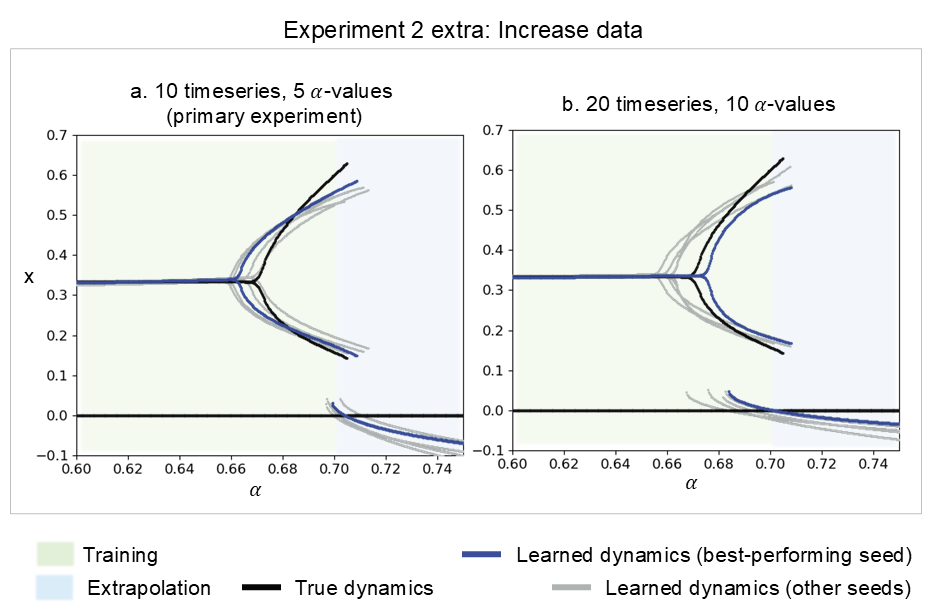}
\caption{\label{fig:extra_experiment} Reconstructed bifurcation diagram for dataset with increased amount of data compared to the primary dataset. Left: primary dataset. Right: bifurcation diagram learned from dataset that included 20 time series for 10 different $\alpha$-values.}
\end{figure}

We also present an additional experiment in which the amount of data was increased to investigate the impact of data availability on model performance. Specifically, we doubled the number of sampled $\alpha$-values in the training set compared to the primary dataset. However, this increase did not lead to a substantial improvement in the model’s accuracy (Fig. \ref{fig:extra_experiment}). The additional samples failed to provide new information for the Neural ODE to further refine its learned dynamics. This suggests that the model could infer no new dynamics that were not yet represented in the primary dataset. These results highlight that, in this setting, simply increasing the amount of data does not necessarily translate to better generalization or accuracy. 

